\documentclass[runningheads]{llncs}
\usepackage{graphicx}
\usepackage{booktabs}
\usepackage[table,xcdraw]{xcolor}
\usepackage{float}
\usepackage[colorlinks=true, allcolors=black]{hyperref}
\restylefloat{table}

\usepackage{etoolbox}
\patchcmd{\bibliography}{\chapter*}{\section*}{}{}
\usepackage[sort, numbers]{natbib}

\usepackage{multicol}

\makeatletter
\newcommand*{\centerfloat}{%
  \parindent \z@
  \leftskip \z@ \@plus 1fil \@minus \textwidth
  \rightskip\leftskip
  \parfillskip \z@skip}
\makeatother

\usepackage{listings}
\usepackage{chngcntr}
\AtBeginDocument{\counterwithin{lstlisting}{section}}
\begin{document}

\title{Knowledge Integration for Disease Characterization:
{A Breast Cancer Example}}

\titlerunning{Knowledge Integration for Breast Cancer Characterization}

\author{Oshani Seneviratne, Sabbir M. Rashid, Shruthi Chari, James P. McCusker, Kristin P. Bennett, James A. Hendler, and Deborah L. McGuinness}

\authorrunning{Seneviratne et al.}

\institute{Rensselaer Polytechnic Institute, Troy NY 12080, USA
}

\maketitle

\begin{abstract}
With the rapid advancements  in cancer research, the information that is useful  for characterizing disease, staging tumors, and creating  treatment and survivorship plans has been changing at a pace that creates challenges when physicians try to remain current.
One example involves increasing usage of biomarkers when characterizing the pathologic prognostic stage of a breast tumor.
We present our semantic technology approach to support cancer characterization and demonstrate it in our end-to-end prototype system that collects the newest breast cancer staging criteria from authoritative oncology manuals to construct an ontology for breast cancer.
Using a tool we developed that utilizes this ontology, physician-facing applications can be used to quickly stage a new patient to support identifying risks, treatment options, and monitoring plans based on authoritative and best practice guidelines.
Physicians can also re-stage existing patients or patient populations, allowing them to find patients whose stage has changed in a given patient cohort.
As new guidelines emerge, using our proposed mechanism, which is grounded by semantic technologies for ingesting new data from staging manuals, we have created an enriched cancer staging ontology that integrates relevant data from several sources with very little human intervention.

\\

\textbf{Resource:} \normalfont{\url{https://cancer-staging-ontology.github.io}}

\end{abstract}

\section{Introduction}

Our goal is to improve health knowledge infrastructures by use of semantic technologies to support data integration in an environment of quickly evolving medical information. 
We present a prototype system that uses semantic technologies to integrate medical information relevant for characterizing breast cancer.
Our system can automatically parse the guidelines from the cancer staging manual and construct OWL axioms \cite{bechhofer2009owl} that can be used to infer recommended personalized options for patients. 
These inferences are made using the data related to the treatment and monitoring of the disease that are represented in RDF \cite{klyne2006resource}.

\subsection{Background}
\label{background}

The authoritative staging system is published by the American Joint Committee on Cancer (AJCC). 
As the inaugural authors of the cancer staging manuals have stated in \cite{beahrs1988manual}: 

\begin{quote}
``Staging of cancer is not an exact science. As new information becomes available about etiology and various methods of diagnosis and treatment, the classification and staging of cancer will change.'' 
\end{quote}

Since the inception of the cancer staging manual in 1977, there have been eight editions. The latest AJCC Cancer Staging Manual, Eighth Edition (AJCC $8^{th}$ Edition)\cite{amin2017eighth}, makes a tangible effort to incorporate biologic and molecular markers to create a more contemporary personalized approach using pathologic prognostic staging. 
This has increased the complexity of the staging criteria.

In order to stage tumors, many physicians rely on cancer staging manuals, or compact `cheat sheets' derived from the contents of these manuals.
However, since the new staging guideline incorporates additional data streams, the physicians have to traverse increasingly tedious decision trees. 

In terms of discovering relevant treatment and monitoring options based on the stage, or more broadly the characterization of the disease, physicians usually refer to the National Comprehensive Cancer Network (NCCN) Guidelines \cite{kim2013nccn}. 
Navigating these guidelines also is often a tedious process.
Furthermore, in order to keep up with the growing and rapidly changing body of knowledge, physicians may also use subscription services such as \emph{UpToDate}\footnote{\emph{UpToDate} -  a clinical decision support resource: \url{http://www.uptodate.com}}, which has articles on many of the state of the art topics in medicine, including cancer.
However, physicians may not have enough time to sift through these articles and ascertain the information that is relevant for the case at hand.

\subsection{Related Work}
\label{related-work}

Initial work related to an ontology that captured cancer staging information is  available in Massicano et al. \cite{massicano2015ontology} for the AJCC $6^{th}$ edition \cite{singletary2003classification}.
Boeker et al. \cite{boeker2016tnm} have also created an ontology for the same guideline in which they focus on tumors in the colon and rectum.
The biggest difference between the previous ontologies and our cancer staging ontology is the inclusion of additional biomarkers as per the AJCC $8^{th}$ edition staging criteria, which were not available in the previous staging editions. These biomarkers used in the new edition significantly increased the complexity of the criteria required to stage a tumor.
Additionally, the previous ontologies do not model real-world representations of the tumor concepts in their axioms nor specify those in the comments.
In those ontologies, the tumor is of a certain \textit{rdf:type T} (class representing severity of tumor size: \textit{T0}-\textit{T4}), \textit{N} (class representing the severity of the spread to the lymph nodes: \textit{N0}-\textit{N3}) and \textit{M} (class representing whether the cancer has metastasized: \textit{M0}-\textit{M1}). In the real world, representation for tumor size T has a value in millimeters (or centimeters) that is used to derive a T value of 0-4. Similarly, N has a value for the number of lymph nodes affected that is used to derive the severity rating from 0-3. 
Thus, their approach of representing the cancer characterization using just the \textit{rdf:type} to the corresponding T, N, M classes is problematic because when any of these derived classifications change as per a new guideline, the RDF graph has to change with it, representing the new classification. In our knowledge graph, these values are encoded as attributes to give them temporal extent, avoiding this problem.

Furthermore, in addition to including classes for all cancer stages for the respective guideline, we also map the breast cancer terms to community-accepted terms from the National Cancer Institute thesaurus (NCIt)~\cite{golbeck2011national}, and incorporate recommended tests and treatment plans from the openly reusable Clinical Interpretations of Variants in Cancer (CIViC) \cite{griffith2017civic} data that can be used to provide stage specific recommendations.
Furthermore, our ontology includes terms
that are not included in NCIt or AJCC, such as more specific subclasses of tumor characteristics (\textit{T1}, \textit{T1\_as}, \textit{T1\_am}, \textit{T1NOS}, etc.) that are available in the Surveillance, Epidemiology, and End Results (SEER) dataset~\cite{hayat2007cancer}.

\subsection{Overview of the Knowledge Integration System for Breast Cancer Characterization}

\begin{figure*}
\centering
\includegraphics[width=0.93\textwidth]{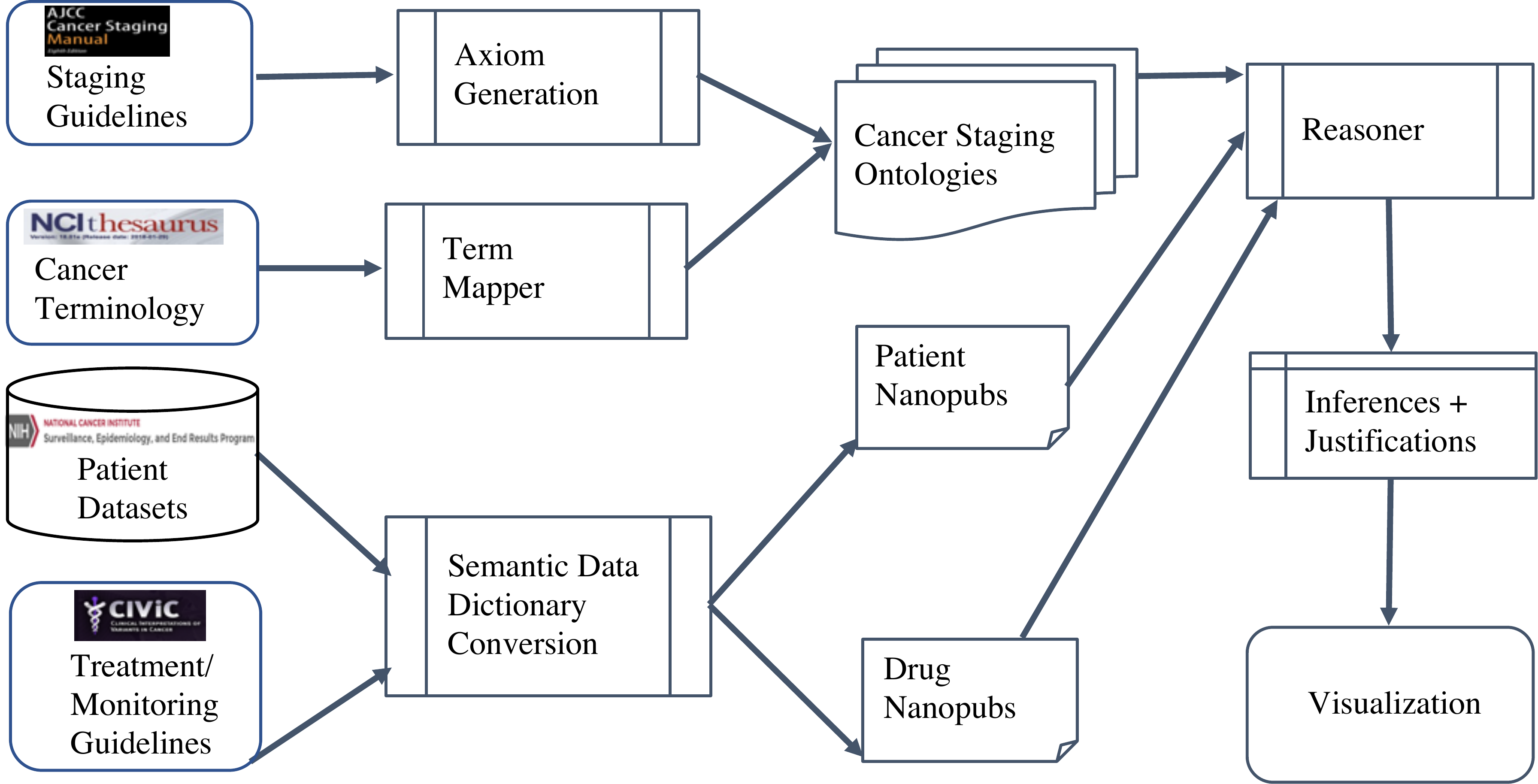}
\caption{\label{fig:architecture}Knowledge Integration Architecture for Breast Cancer Characterization}
\end{figure*}

We developed our prototype primarily to address the issue of rapidly changing information in characterizing disease, specifically breast cancer.
Since manual look-up of the breast cancer staging criteria is prone to human error, our system was designed to support automated navigation through the tedious decision trees to minimize any look up errors.  We also provide support for integration of data from various sources. 
Fig. \ref{fig:architecture} depicts the overall knowledge integration architecture that will be explained in detail in the following sections. 

\section{Development of the Cancer Staging Ontologies}
\label{ontology-creation}

As mentioned in related work (Section \ref{related-work}) the last known staging ontologies were created for the AJCC $6^{th}$ edition. 
There are no ontologies for the AJCC $7^{th}$ and $8^{th}$ editions to the best of our knowledge. 
We describe the process we followed when constructing these new staging ontologies, accounting for the complexity of the data streams in the new guideline.

\subsection{Cancer Staging Terms}
\label{cst}

The previous breast cancer staging guidelines (i.e. AJCC $7^{th}$ edition \cite{edge2010american} and earlier) only considered anatomical features such as the size of the tumor (T), the number of lymph nodes affected (N), and  whether the cancer has metastasized (M).
Additionally considered in the new staging guidelines \cite{amin2017eighth} are biomarkers including human epidermal growth factor receptor 2 (HER2), estrogen (ER) and progesterone (PR) receptor statuses and tumor grade (Grade). 
This addition has led to a more complex set of rules for staging criteria using the classes corresponding to the specific stages in the AJCC $7^{th}$ and $8^{th}$ editions that we incorporated into our Cancer Staging Terms (CST) ontology. 

\begin{figure*}
\centering
\includegraphics[width=1.1\textwidth]{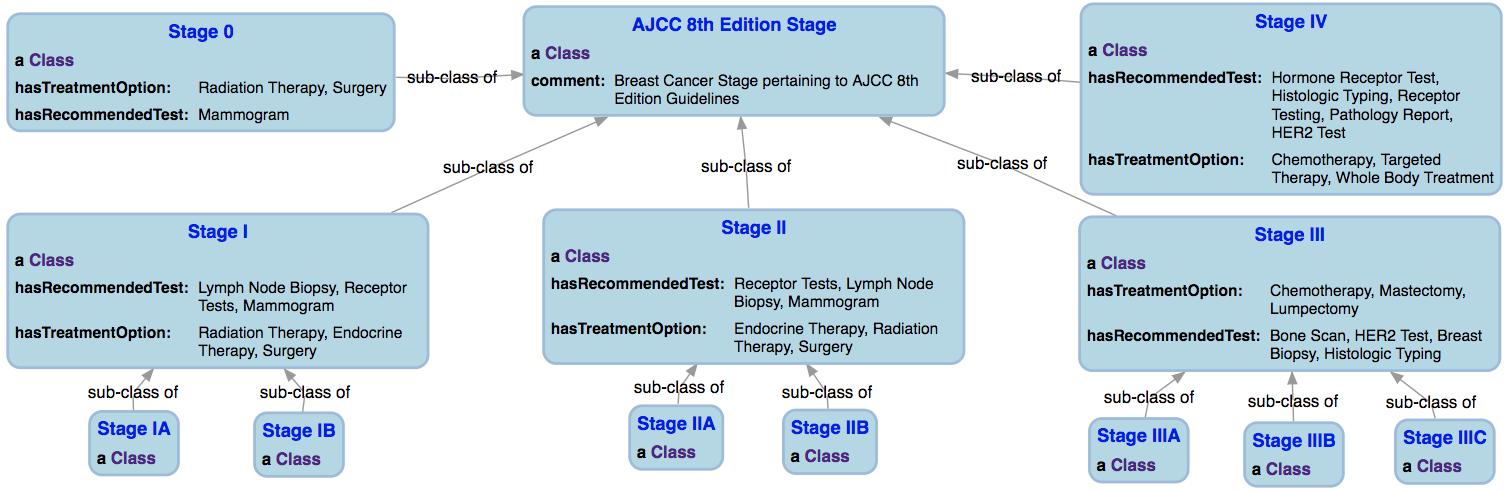}
\caption{\label{fig:stage_hierarchy}Stage Hierarchy of the AJCC Cancer Staging $8^{th}$ Edition}
\end{figure*}

Fig. \ref{fig:stage_hierarchy} depicts the $8^{th}$ staging edition staging class hierarchy. 
Each stage class includes the properties \emph{cst:hasRecommendedTest}, \emph{cst:hasTreatmentOption}, and \emph{rdfs:subClassOf} assertions where applicable. We added the \emph{rdfs:comments} to better describe the concepts in the ontology based on the descriptions available in the medical literature and to support explanation. 

\begin{figure*}
\centering
\includegraphics[width=1.1\textwidth]{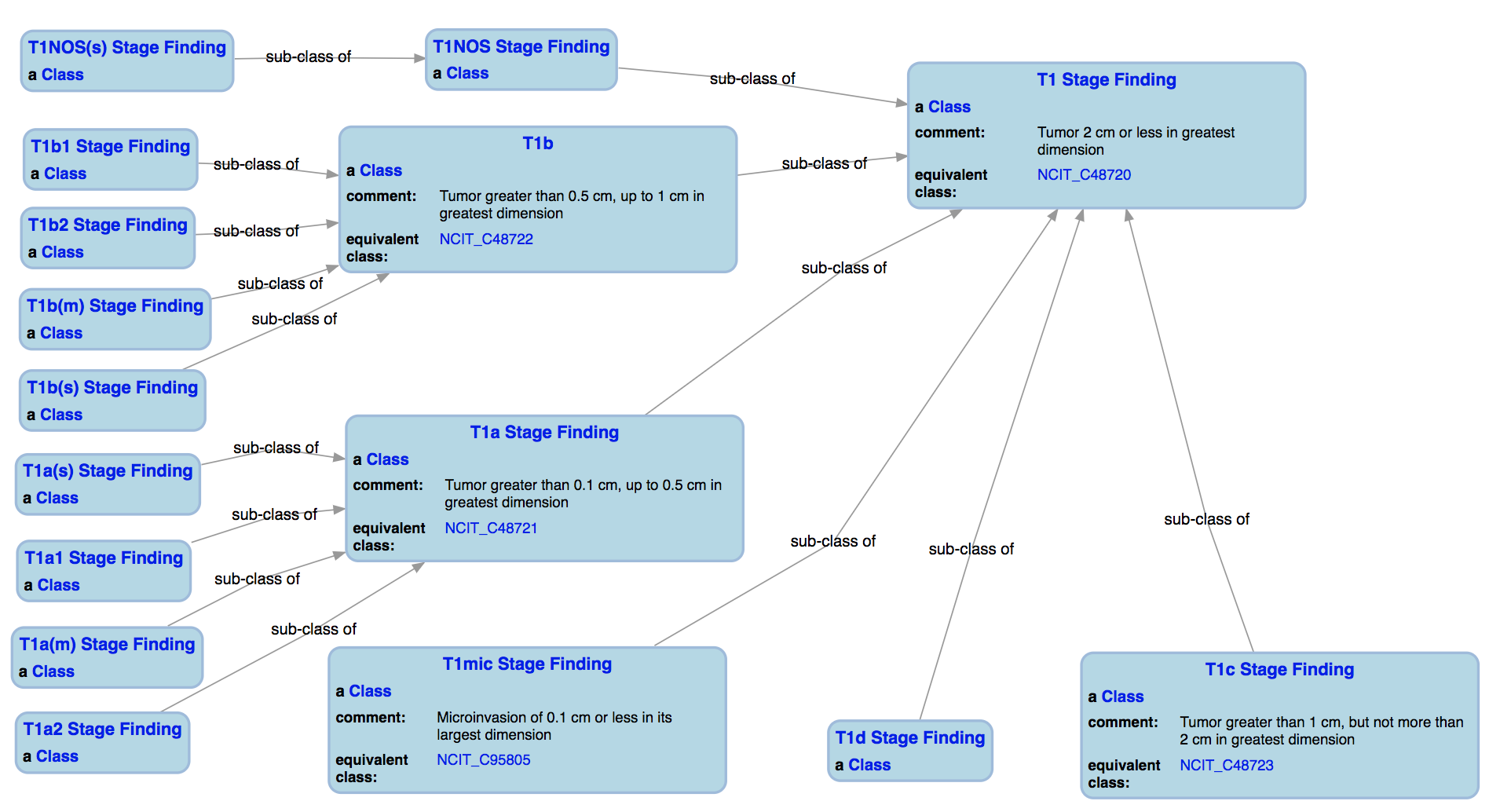}
\caption{\label{fig:t_hierarchy}Hierarchy of the Tumor Size (T) Classes in Our Integrated Ontology. \emph{We created the tumor size classes in the left two columns to support our integration and reasoning. These classes reflect content in SEER~\cite{hayat2007cancer} and not AJCC~\cite{amin2017eighth}.}}
\end{figure*}

Furthermore, in the AJCC staging manuals, and in the data we ingested from other sources, we observed different subclasses for the broader classification of the features considered, i.e. T, N, M, HER2, ER, PR, and Grade in the ontology.  
Fig. \ref{fig:t_hierarchy} depicts a small subset of these classifications, which includes various Tumor size (T) classes. Similarly, there are other subclass assertions, and mappings to the NCIt classes for N, M, HER2, ER, PR, and Grade. 
We augmented these classes with the \textit{rdfs:comment}, \textit{rdfs:label}s, and the \textit{owl:equivalentClass} obtained from NCIt \cite{golbeck2011national}. 
These \textit{rdfs:comment}s and \textit{rdfs:label}s are used to explain a particular conclusion resulting from the application of a reasoner utilizing the ontology explained in detail in Section \ref{sec:inferencer}. 

\subsection{Translating Staging Criteria into Structured Mappings}
\label{manual-to-mappings}

We extracted 19 criteria from AJCC $7^{th}$ edition, and 407 criteria for clinical prognostic stage grouping from AJCC $8^{th}$ edition. A script was necessary for the $8^{th}$ edition since the complexity of the staging guideline has increased with the addition of the biomarkers. Table \ref{tbl-combinations} illustrates the number of different combinations for staging criteria observed in the two staging guidelines. The non-linear expansion of the number of combinations is due to the complex interaction of the additional biomarkers HER2, ER, PR and Tumor Grade. 

\begin{table}[H]
\centering
\caption{Number of Feature Combinations for Determining Stage}
\label{tbl-combinations}
\begin{tabular}{|l|l|l|l|l|l|l|l|l|l|}
\hline
\textit{\textbf{Stage}}           & \textit{\textbf{0}} & \textit{\textbf{IA}} & \textit{\textbf{IB}} & \textit{\textbf{IIA}} & \textit{\textbf{IIB}} & \textit{\textbf{IIIA}} & \textit{\textbf{IIIB}} & \textit{\textbf{IIIC}} & \textit{\textbf{IV}} \\ \hline
\textbf{AJCC $7^{th}$ Edition} & 1                   & 1                    & 2                    & 3                     & 2                     & 5                      & 3                      & 1                      & 1                    \\ \hline
\textbf{AJCC $8^{th}$ Edition} & 1                   & 57                   & 33                   & 77                    & 39                    & 82                     & 92                     & 25                     & 1                    \\ \hline
\end{tabular}
\end{table}

For each of the two staging guidelines, we created corresponding `map files' to represent the conditions required for a tumor to be classified a certain stage from 0-IV. 
We created 18 such map files for the two guidelines (AJCC $7^{th}$ and $8^{th}$ editions), with 9 map files representing each stage from 0, IA, IB, IIA, IIB, IIIA, IIIB, IIIC, and IV. Each line in the map file in a $7^{th}$ edition stage contains the set of possible T, N and M combinations that would result in that stage being assigned to the tumor. The map files for the $8^{th}$ edition followed a similar form, but also included the additional features HER2, ER, PR and Grade. 

If any of the features can be \textbf{\emph{any}} value for a tumor to be staged, the map file omitted those corresponding features, and only included the features that mattered. For example, in order for a tumor to be classified stage IV in both the guidelines, the only criteria necessary was the `M' (whether the cancer has metastasized) to be true. Regardless of any other combinations of the other features T, N in the $7^{th}$ edition, and additionally HER2, ER, PR and Grade in the $8^{th}$ edition, the tumor will always be classified stage IV, thus only one combination is available for both the guidelines for determining stage IV.

\subsection{Structured Mappings to Ontology}

In order to automatically generate OWL axioms for the staging criteria, we utilized the map files created in Section \ref{manual-to-mappings}. These map files were parsed using a script, where the property \textit{owl:intersectionOf} was leveraged in creating the axioms. For example, in order for a tumor to be classified as Stage IA in the AJCC $7^{th}$ edition (i.e. ${AJCC7\_Stage\_IA}$), a tumor profile must satisfy the axiom in Listing \ref{AJCC7_IA}. However, for the same tumor to be classified as Stage IA in the AJCC $8^{th}$ edition (i.e. ${AJCC8\_Stage\_IA}$), only one of the 57 axioms must be satisfied (Listing \ref{AJCC8_IA} demonstrates one such axiom). We developed the breast cancer staging ontology for the AJCC $7^{th}$ edition (BCS7) and the ontology for the AJCC $8^{th}$ edition (BCS8) using the above-mentioned procedure to codify all the axioms related to classifying tumors.

\begin{lstlisting}[caption=The Only OWL Axiom for a Tumor to be Classified as \textit{Stage IA} in the AJCC $7^{th}$ Edition, label=AJCC7_IA]
@prefix cst: <http://idea.tw.rpi.edu/cancer_staging_terms.owl#> .
[] a owl:Class; rdfs:subClassOf cst:AJCC7_Stage_IA;
    owl:intersectionOf ( cst:T1 cst:N0 cst:M0 ).
\end{lstlisting}

\begin{lstlisting}[caption=One of the Many OWL Axioms for a Tumor to be Classified as \textit{Stage IA} in the AJCC $8^{th}$ Edition, label=AJCC8_IA]
[] a owl:Class; rdfs:subClassOf cst:AJCC8_Stage_IA;
    owl:intersectionOf ( cst:T1 cst:N0 cst:M0 cst:Grade1 
    cst:HER2_Neg cst:ER_Neg cst:PR_Pos ).
...
\end{lstlisting}

\section{Integrated Cancer Knowledge Graph}
\label{cancer-knowledge-graph}

We chose RDF \cite{klyne2006resource} as the underlying knowledge representation model to handle heterogeneous data while providing interoperable representations.
The CST, BCS7 and BCS8 developed in Section \ref{ontology-creation}, are part of our Integrated Cancer Knowledge Graph. Additionally, we extracted data from crowd sourced, open source, reusable cancer resources to augment the knowledge graph with treatment and monitoring options based on the stage inferred using the cancer staging ontologies developed.

\subsection{Integrating Data from Other Cancer Data Sources}

There are many services that provide vast collections of data that may be useful and relevant in a cancer knowledge graph. Some of these services include CIViC \cite{griffith2017civic}, OncoKB \cite{chakravarty2017oncokb}, MyCancerGenome \cite{micheel2014my} and Integrative Onco Genomics \cite{gonzalez2013intogen}.

As a proof of concept, we incorporated data from CIViC \cite{griffith2017civic}, which has crowd sourced, open source and reusable data that identifies drugs that may interact with biomarkers. Additionally from the CIViC data dumps, related articles and their trust ratings captured in the form of provenance were also incorporated.
Nanopublications \cite{mons2009nano} were created for this data using a semantic annotation approach called Semantic Data Dictionaries (SDDs) \cite{rashidsemantic}, which simplifies the ability to express the full semantics of a dataset. The SDD process \cite{rashidsemantic} allowed us to link the data concepts with each other, as well as reference implicit entities in the data, and link the corresponding data elements as characteristics of these entities.
The concepts contained in the data records needed to be mapped with related terms from domain specific ontologies such as NCI thesaurus (NCIt) \cite{golbeck2011national} and Uniprot \cite{uniprot2014uniprot}, as well as general purpose ontologies such as Semanticscience Integrated Ontology (SIO) \cite{dumontier2014semanticscience}.

The dictionary mapping table of the SDD that was used for CIViC maps 14 different features in the dataset such as \textit{Drugs}, \textit{Status}, \textit{Evidence ID}, \textit{Evidence Level}, \textit{Gene}, \textit{Variant}, \textit{Disease }and the \textit{Trust Rating} to the respective classes available in SIO and NCIt. These classes are used for type assignment when creating a knowledge graph from the data. For example, the \emph{Drug} column in the dataset is mapped to \emph{sio:Drug}, \emph{Gene} column to \emph{sio:Gene}, etc. Furthermore, the classes specified in the \emph{attributeOf}, \emph{inRelationTo} and \emph{wasDerivedFrom} are used in semantically modeling relationships in the generated nanopublications. 

A codebook was used to map over 200 specific values found in the CIViC data to the corresponding terms in existing ontologies. The \textit{Disease} types that were found in CIViC were mapped to concepts in the Human Disease Ontology (DOID) \cite{schriml2011disease}, Experimental Factor Ontology (EFO) \cite{malone2010modeling}, and NCIt \cite{golbeck2011national}. For example, the concept for the \emph{HER2-receptor Positive Breast Cancer} in our knowledge graph is mapped to concepts such as \emph{efo:1000294, doid:0060079, ncit:C53556}\footnote{These specific mappings were looked up using Ontobee (\url{http://www.ontobee.org}).}.
Similarly, drugs were mapped to concepts from the Drug Bank \cite{wishart2006drugbank}, the Drug Ontology (DRON) \cite{hanna2013building}, and/or Chemical Entities of Biological Interest (ChEBI) Ontology \cite{degtyarenko2007chebi}, genes were mapped to terms in Uniprot \cite{uniprot2014uniprot}, etc.

\section{Converting Patient Records to RDF}
\label{seer}

In order to evaluate our cancer staging ontology, we needed cancer patient data that included the characteristics of the tumor in RDF, ideally in the nanopublications format \cite{mons2009nano}.
The SEER datasets \cite{hayat2007cancer} contained the desired data which included demographic information, tumor stage as per the older AJCC $6^{th}$ edition, and the survival status of patients treated from 1980-2012.
We browsed the datasets using the statistical software, SEER*Stat\footnote{SEER*Stat: https://seer.cancer.gov/seerstat}, and downloaded a subset of the data to create the patient nanopublications.

Due to the anonymity and privacy constraints on the medical data, the SEER patient records lacked any identifying information like the patient name.
However, for our use case, i.e. to model a patient, we needed an identifying attribute, so we annotated the patient records with names from Python's Natural Language Toolkit (NLTK) name corpus \cite{bird2004nltk} to assign a name to each patient record.
The patient data was then fed through the SDD pipeline \cite{rashidsemantic} to generate knowledge graphs that included nanopublications that captured the attributes of a patient and where that information came from within an assertion in the patient graph. 
Utilization of the SDD approach allowed us to semantically represent relationships such as the age of patient at diagnosis (i.e. the attribute \textit{sio:Age} as \textit{sio:attributeOf} the patient which \textit{sio:existsAt} the time of diagnosis). We mapped 29 such features for a patient record in the SEER dataset in the data dictionary, and the codebook contains 100+ mappings to terms in NCIt. 

Since some of the values occurring in SEER did not match existing terms in the ontologies, 
we leveraged our Cancer Staging Terms (CST) ontology, introduced in Section \ref{ontology-creation}.
A codebook mapping corresponding to SEER was defined that would generate standard values and map commonly occurring terms to their ontology equivalents.
As the structured format of the data is insufficient to capture the implicit linkages within the attributes of the dataset, a SEER dictionary mapping was defined that established the entity-attribute mappings to facilitate the conversion of the data to the named graphs with nanopublications.

\section{Inference Agent}
\label{sec:inferencer}

We developed a deductive inference agent on the Whyis knowledge graph framework \cite{whyis} to infer the stage of a tumor, and the corresponding treatment/monitoring plans. 
Whyis provides an environment for automated generalized inference over changes to the knowledge graph, supporting the generation of derived knowledge.
The framework enables knowledge curation using a Semantic Extract, Transform, and Load tool for creating RDF from tabular sources, as well as automated mapping of external linked data knowledge sources. 
Furthermore, developers can create custom views for visualizing the data in the underlying knowledge graph.

The Whyis inference agent was built to reason over the nanopublications pertaining to the patient data records constructed in Section \ref{seer}, using the cancer staging ontologies CST, BCS7 and BCS8 introduced in Section \ref{ontology-creation}. 
While the SDD process \cite{rashidsemantic} allowed us to  model the data easily, it resulted in some challenges in terms of writing inference rules in OWL, such as finding appropriate paths between entities or attributes specific to the nanopublications, as well as inference over individuals rather than just classes.
To address these issues, we decided to take a route similar to SPARQL DL \cite{sirin2007sparql} and built SPARQL templates for different OWL reasoning profiles, as well as custom inference rules based on the SDD files, to be consumed by the inference agent. 

An example configuration for an OWL inference rule is shown in Listing \ref{owl-inference-rule}, and an example custom rule, auto-generated with the utilization of files generated by the SDD process is shown in Listing \ref{custom-inference-rule}.

\begin{lstlisting}[caption=Example Configuration for an OWL Inference Rule (Class Subsumption)), label=owl-inference-rule]
"Class Subsumption Closure": (
    where = "?resource rdfs:subClassOf ?class .
    ?class rdfs:subClassOf+ ?superClass .",
    construct="?resource rdfs:subClassOf ?superClass .",
    explanation="Since {{class}} is a subclass of {{superClass}}, 
    any class that is a subclass of {{class}} is also a subclass 
    of {{superClass}}. Therefore, {{resource}} is a subclass of 	
    {{superClass}}.")
\end{lstlisting}

\begin{lstlisting}[caption=Example Configuration for a Custom Inference Rule (One of the Criteria for a Tumor to be Classified as Stage IIIA in the AJCC $8^{th}$ Edition), label=custom-inference-rule]
"AJCC8 Stage IIIA": (
    resource="?Tumor",
    prefixes="...",
    construct="?Tumor cst:hasAJCCStage cst:AJCC8_Stage_IIIA .",
    where=tnm_where + 
      ?T rdf:type cst:T3 .
      ?N rdf:type cst:N3 .
      ?M rdf:type cst:M0 .
      ?Grade rdf:type cst:Grade1 .
      ?HER2 rdf:type cst:HER2_Pos .
      ?ER rdf:type cst:ER_Pos .
      ?PR rdf:type cst:PR_Pos .
\end{lstlisting}

These configurations are used to instantiate the following variables in the inference agent: \texttt{resource}, \texttt{prefixes}, \texttt{where}, \texttt{construct} and \texttt{explanation}.
The \texttt{prefixes} and \texttt{where} variables are used in a SPARQL query that selects relevant URIs from the triple store.
The \texttt{resource} variable is used to refer to which element returned by the query will be appended new triples. 
The form of the new triples that will be added is specified in the \texttt{construct} variable.
An explanation for the rule creating this new knowledge is specified in the \texttt{explanation} variable. 

\subsection*{Generating Explanations}
Our data conversion process captures the provenance of the various sources, which we convert to nanopublications, as well as the explanations behind why specific assertions were inferred. These natural language explanations make it easier for a non-technical user, who might not have an in-depth knowledge of the staging rules, to understand why a certain stage was inferred.
When an explicit \texttt{explanation} is not provided in the rule, it is derived from the \texttt{where} clause used to create the assertion corresponding to the inference. 
The \texttt{explanation} is then associated with that assertion on the new inferred stage using the \textit{prov:used} property. As an example, when the custom inference rule specified in Listing \ref{custom-inference-rule} is fired on \emph{Patient D}, whose tumor satisfies the criteria given in the \texttt{where} clause in that rule, an explanation similar to the one shown in Listing \ref{lst-justification} will be generated. 
For better readability of the explanation, the \textit{rdfs:label} or \textit{rdfs:comment} of the values that get bound to variables such as \texttt{?T,?N,?M}, etc. (i.e. `Primary Tumor size', `Degree of spread to lymph nodes', `Presence of distant metastasis', etc.) are used instead of the actual class names.

\begin{lstlisting}[caption=Example of Explanation for Inferring a Stage in the AJCC $8^{th}$ Edition, label=lst-justification]
Patient D's tumor was found to be AJCC8 Stage IIIA since the 
following are true:
- Primary Tumor size is T3 .
- Degree of spread to lymph nodes is N3 . 
- Presence of distant metastasis is M0 .
- Tumor Grade is Grade3 .
- Human Epidermal growth factor Receptor 2 (HER2) is Positive.
- Estrogen Receptor (ER) is Positive.
- Progesterone Receptor (PR) is Positive.
\end{lstlisting}

Using a similar strategy, we are able to identify possible drug treatment plans from the cancer database CIViC by equating the disease type that the drugs target, to the inferred cancer stage of the patient. To achieve this, we generated custom inference rules from the CIViC SDD files, and once the inferencer runs these rules on the patient nanopublications, the corresponding explanations were generated and attached to the stage assertion nanopublications.

\section{Visualization of the Cancer Characterization}
\label{whyis-view}

\begin{figure}
\centering
\begin{minipage}[b]{0.49\textwidth}
\includegraphics[width=1.05\textwidth]{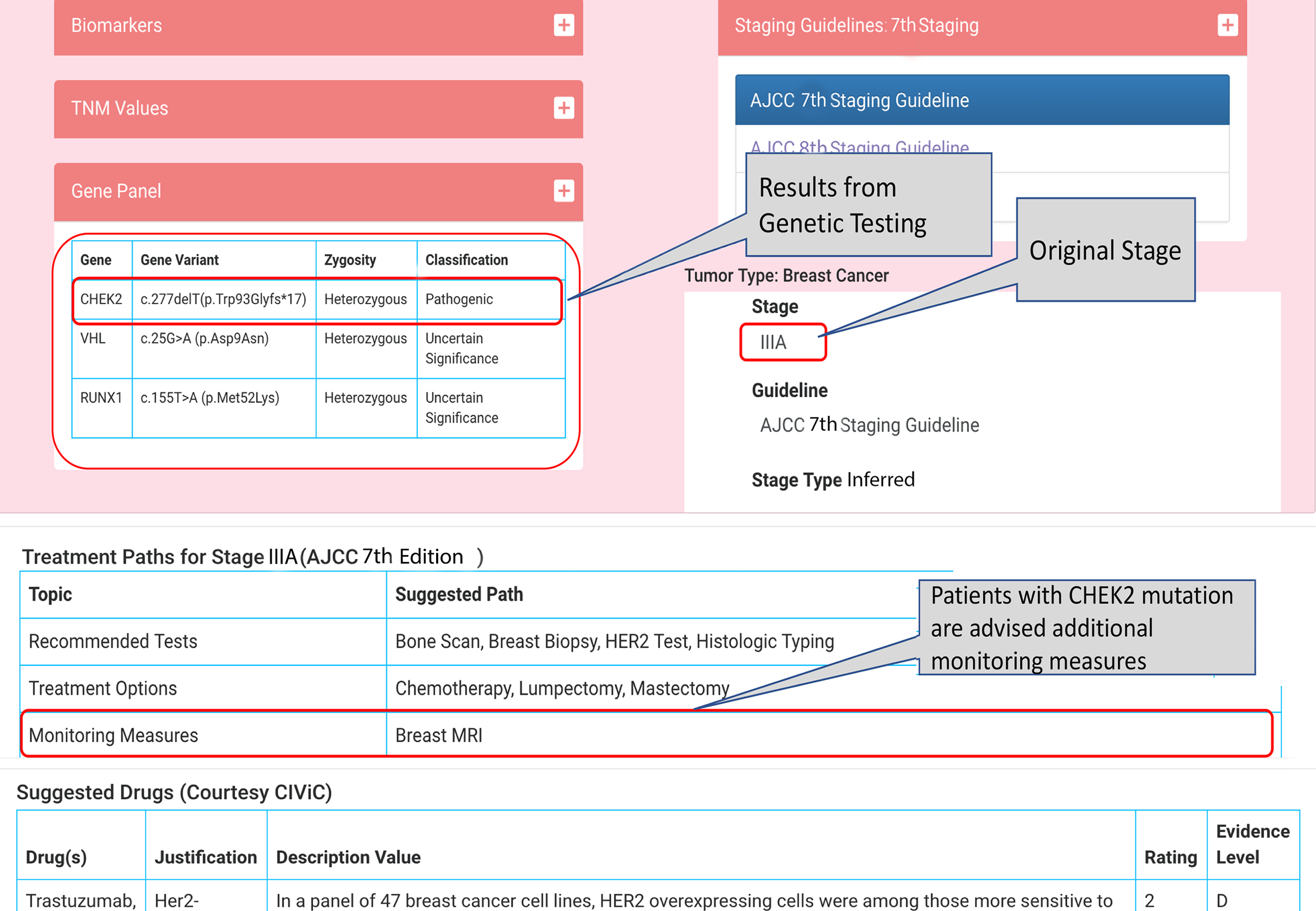}
\caption{\label{fig:6th} AJCC $7^{th}$ Edition Staging Characterization}
\end{minipage}
\hfill
\begin{minipage}[b]{0.49\textwidth}
\includegraphics[width=1.075\textwidth]{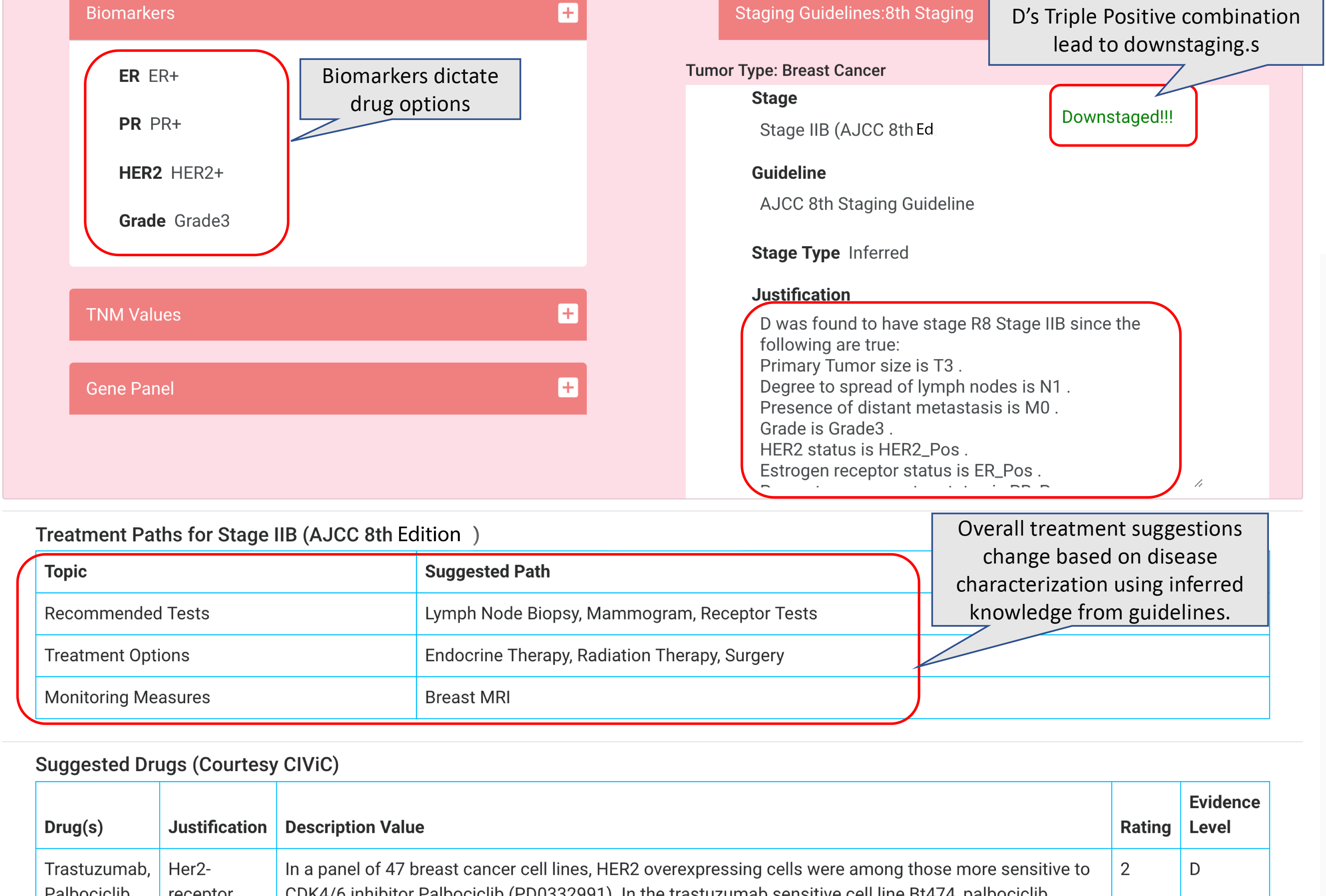}
\caption{\label{fig:8th}AJCC $8^{th}$ Edition Staging Characterization}
\end{minipage}
\end{figure}

In order to demonstrate the integrated cancer knowledge graph and the reasoning capabilities of the Whyis inference agent, we built a visualization tool that displays different treatment paths and guideline impacts to a patient in the form of interactive reports as introduced by Kennedy et al. \cite{kennedy1999interactive}. 
The visualization is built on the Whyis knowledge graph framework (introduced in Section \ref{sec:inferencer}).  
When a user, say a physician, selects a patient record, they are presented with information that helps enhance their diagnostic process, and in some cases, eliminates the manual labor of walking through the decision trees in the guidelines to support cancer staging decisions.
As can be seen in the Fig. \ref{fig:6th} and Fig. \ref{fig:8th}, the view is divided into four sections: (1) \emph{Patient Details}, (2) \emph{Biomarker and Staging}, (3) \emph{Treatment Plan}, and (4) \emph{Suggested Drugs}.

In this visualization tool, it is possible to choose between the three latest AJCC staging guidelines, i.e. AJCC $6^{th}$, $7^{th}$, and $8^{th}$ editions. Once a guideline is selected, the view dynamically loads newly derived knowledge using asynchronous JavaScript SPARQL POST requests. The derived knowledge includes the inferred stage, whether this is an up-stage/down-stage/no-change, and the explanations behind the inferred stage. Based on the inferred stage for the guideline selected, the corresponding treatment and monitoring options available in our integrated cancer knowledge graph (i.e. CIViC drug nanopublication records) are also queried and presented to the user. 

A screenshot of a patient's report as per the older $7^{th}$ edition is shown in Fig. \ref{fig:6th}, and the same patient's report according to the newer $8^{th}$ edition is shown in Fig. \ref{fig:8th}. Note the differences in the inferred stage--the patient is down-staged from IIIA in the $7^{th}$ edition to IIB in the $8^{th}$ edition. There are also some changes to the treatment and monitoring options based on this new inferred stage.

\section{Evaluation}

\begin{figure*}
\centering
\includegraphics[width=0.9\textwidth]{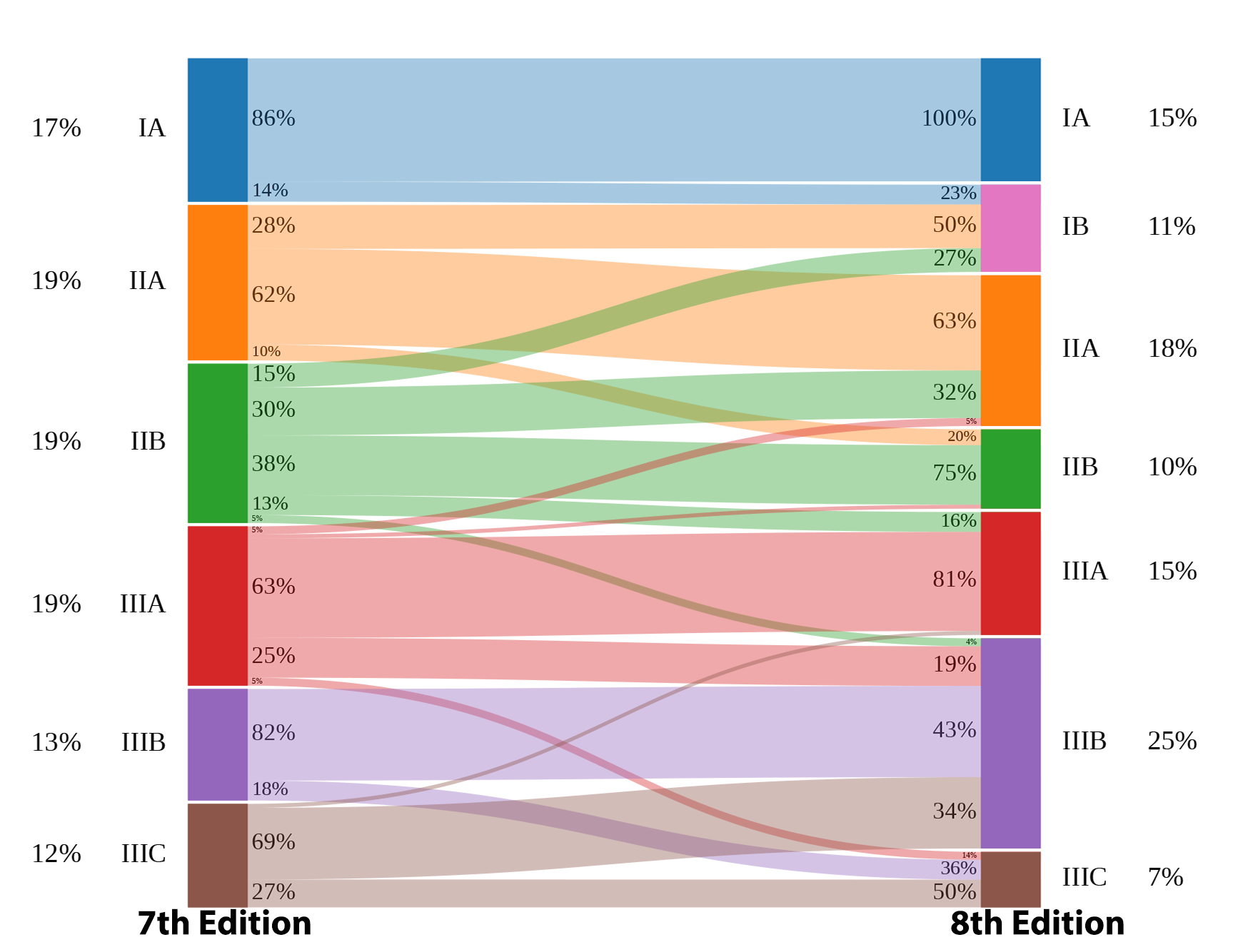}
\caption{\label{fig:transition_matrix}Stage Transitions of 
$250$ Patient Records from SEER}
\end{figure*}

We used our cancer staging ontologies and the inference agent on 250 randomly selected SEER patient records to estimate prevalence of stage changes between different staging guidelines.
We anticipated a number of changes given that the latest AJCC $8^{th}$ edition utilizes additional biomarkers to determine stage. 
These SEER patient records were first transformed into nanopublications using the SDD process~\cite{rashidsemantic} as explained in Section \ref{seer}, after which our inference agent was applied to determine the stage as per the two guidelines. 

The aggregated view of these stage transitions from the AJCC $7^{th}$ to the $8^{th}$ edition is shown in Fig. \ref{fig:transition_matrix}. 
As can be seen in the figure, a majority of the patients' stage did not change, but a statistically significant percentage of patients were either up-staged or down-staged. 
For example, out of the patients who were assigned to have stage IIB cancer according to the $7^{th}$ edition ($19\%$ of the population), 15\% were down-staged to IB, 30\% were reclassified to  IIA, 38\% remained in stage IIB, and 13\% and 5\% were up-staged to IIIA and IIIB respectively. 

This indicates that there is a strong need for re-characterizing breast cancer according to the new guideline. Our ontologies and the supporting tools provide the first step in this process.

\section{Discussion}
\label{discussion}

We have utilized semantic technologies for all aspects of our system: from characterizing breast cancer and representing synthetic patient data to loading structured and unstructured treatment and monitoring data into a knowledge graph. 

For the integrated cancer knowledge graph generation, we mapped concepts in several datasets using a codebook and modeled a structure amongst the attributes using the dictionary mapping table. 
The deductive inference agent we developed leverages SPARQL DL reasoning, where queries are used to select existing triples and construct new triples. 
This was done over common inference rules including class subsumption and class or property equivalence closures, as well as custom rules pertaining to the cancer staging. 
The inference was applied to heterogeneous data sources in our integrated cancer knowledge graph to seamlessly derive new knowledge by applying the inference rules.
The visualization we created is able to react to changes in the triple store that results in automatic updates to what the user is seeing. 
The information in our cancer knowledge graph is kept current with  periodic semantic extract transform load updates.
Our system allows one to consider a multitude of parameters related to tumor biology as well as standard pathology simultaneously and can easily updated to support new classification criteria.

\subsection*{Resource Contributions}
\label{sec:contributions}

We expect the following publicly available artifacts, along with the applicable documentation, 
to be useful resources for anyone interested in analyzing breast cancer data according to the new and the old cancer staging guidelines. 

\begin{multicols}{2}
\begin{enumerate}    
	\item Ontologies:
  		\begin{enumerate}
    		\item Cancer Staging Terms (CST)
    		\item Breast Cancer Staging Ontology for the AJCC $7^{th}$ Edition (BCS7)
			\item Breast Cancer Staging Ontology for the AJCC $8^{th}$ Edition (BCS8)\\
  		\end{enumerate}
    \item Semantic Annotations:
    	\begin{enumerate}
    		\item Semantic Data Dictionaries
    		\item Code Books
            \\ \emph{(for SEER and CIViC)}
  		\end{enumerate}
    \item Source Code:
    	\begin{enumerate}
    		\item AJCC Guideline Extractor
    		\item OWL Axiom Generator 
			\item Whyis Inference Agent
            \item Custom Inference Rules
            \item Visualization\\\\
  		\end{enumerate}
    \item Data:
    	\begin{enumerate}
    		\item SEER Nanopublications
    		\item CIViC Nanopublications
  		\end{enumerate}
    
\end{enumerate}
\end{multicols}

\section{Future Work}

There are many online resources with rapidly changing information from clinical trials, as well as data from basic science research with useful cancer data that can be leveraged to augment the cancer knowledge graph. However, when multiple data streams are combined, especially drug information, there may be inconsistent or ambiguous information. Therefore, we will need to resolve such issues using a combination of provenance, data integrity, and trust in the source and/or the methodology.

The inference agent we developed can be used to identify treatment paths based on a patient's cancer stage. However, the CIViC data~\cite{griffith2017civic} we used for this purpose defines treatment paths for the broader stages (i.e. stage II as opposed to AJCC's narrower IIA or IIB stages). Therefore, we plan to ascertain the correct treatment paths for all the narrower stages and add those in to the cancer knowledge graph. 
We plan to incorporate additional data sources such as the NCCN clinical practice guidelines in oncology \cite{kim2013nccn}, which is the authoritative source for physicians in identifying suitable cancer treatment and monitoring plans. This will allow our inference agent to output the precise treatment paths, in addition to the ones that are obtained by linking the patient's inferred narrower AJCC tumor stage to the broader stage with ontological properties such as \textit{rdfs:subClassOf} relationships.
We also expect the future ontologies to be built using the AJCC API as a resource for all valid values on stage permutations. 

We plan to expand the inference capabilities, which are currently restricted to class, instance,  property subsumption, equivalence, and inversion closures, to other techniques that will help derive even more relevant knowledge. 
For example, we believe it would be useful to infer `patients like me' using instance matching and identify alternate treatment paths that have worked in the past for similar patients, and predict response to a treatment path using temporal reasoning.

As new guidelines will infer new staging results, updates or fixes to the patient data or existing guidelines are needed. The Whyis framework provides an effective mechanism to `retire' old inferences and trigger computation of new ones, as long as the nanopublication has the same URI. The framework tracks a nanopublication URI when a new version is added, removing older versions, as well as any inferences that are made on them. We opted to create different classes for the stage based on the guideline, so that we can switch between different guidelines easily. For example, we have \textit{AJCC7\_Stage\_IIA} and \textit{AJCC8\_Stage\_IIA}, as opposed to a generic \textit{Stage\_IIA}. Therefore, encoding the information about which guideline the staging criteria is from, in the provenance assertion for that triple, without having to make that explicit class, and utilizing the provenance information in the inference to determine the stage per the selected guideline is a useful addition. This change requires versioning of assertions in our integrated cancer knowledge graph, and some changes to the custom inference rules.

\section{Conclusion}

We have presented a prototype knowledge integration system that can be used to encapsulate the breadth of information required to characterize disease. 
The specific domain problem we address is characterizing breast cancer, which today is predominantly done by manually looking up cancer staging guidelines. 
In fact, oncology is moving towards adopting the concept of \emph{Precision Oncology}, in which the treatment plans and therapies are driven by data from personalized genetic markers independent of cancer type \cite{garraway2013precision}. In the future, new guidelines for cancer staging are expected to incorporate genomic test results analyzed in the context of the patient's history, which will further increase the complexity of the staging criteria, requiring automated mechanisms similar to the techniques illustrated in this paper. Therefore, it is our expectation that the resources contributed in this paper and the methodologies to ingest rapidly changing information, will be useful to application designers who are aiming to support next generation precision medicine assistant tools.

\section*{Acknowledgements}
This work is partially supported by IBM Research AI through the AI Horizons Network. We thank our colleagues from IBM (Amar Das, Ching-Hua Chen) and RPI (John Erickson, Alexander New, Rebecca Cowan) who provided insight and expertise that greatly assisted the research.

\bibliographystyle{splncs04}
\bibliography{references}

\end{document}